\documentclass[sigconf]{acmart}

\usepackage{amsmath,amssymb,amsfonts}
\usepackage{algorithm, algorithmic}
\usepackage{graphicx}
\usepackage{textcomp}
\usepackage{xcolor}
\usepackage{comment}
\usepackage{float}
\usepackage{geometry}
\usepackage{balance}

\usepackage[export]{adjustbox}
\usepackage[caption=false, font=footnotesize]{subfig}
\def\BibTeX{{\rm B\kern-.05em{\sc i\kern-.025em b}\kern-.08emT\kern-.1667em\lower.7ex\hbox{E}\kern-.125emX}}

\begin{document}
\fancyhead{}

\title{User-assisted Video Reflection Removal}

\author{Amgad Ahmed}
\affiliation{\institution{School of Computing Science \\ Simon Fraser University}}

\author{Suhong Kim}
\affiliation{\institution{School of Computing Science \\  Simon Fraser University}}

\author{Mohamed Elgharib}
\affiliation{\institution{Max Planck Institute for Informatics}}

\author{Mohamed Hefeeda}
\affiliation{\institution{School of Computing Science \\ Simon Fraser University}}

\begin{abstract}
Reflections in videos are obstructions that often occur when videos are taken behind reflective surfaces like glass. These reflections reduce the quality of such videos, lead to information loss and degrade the accuracy of many computer vision algorithms. A video containing reflections is a combination of background and reflection layers. Thus, reflection removal is equivalent to decomposing the video into two layers. This, however, is a challenging and ill-posed problem as there is an infinite number of valid decompositions. To address this problem, we propose a user-assisted method for video reflection removal. We rely on both spatial and temporal information and utilize sparse user hints to help improve separation. The key idea of the proposed method is to use motion cues to separate the background layer from the reflection layer with minimal user assistance. We show that user-assistance significantly improves the layer separation results.  We implement and evaluate the proposed method through quantitative and qualitative results on real and synthetic videos. Our experiments show that the proposed method successfully removes reflection from video sequences, does not introduce visual distortions, and significantly outperforms the state-of-the-art reflection removal methods in the literature. 
\end{abstract}




\maketitle

\section{Introduction} \label{sec:introduction}

With the popularity of digital devices and videography, videos have become one of the most important information carriers. Users of such devices are encountered with video capturing conditions that can be far from optimal. For example, when taking videos behind glass windows inside a building or a car, reflections from indoor objects may obstruct the outdoor scene of interest. These reflections reduce the quality of such videos and decrease the target objects visibility. 

Removing reflections from videos results in clearer and better-quality videos, which is 
important for professional photographers as well as normal users. Moreover, cameras in self-driving cars are often mounted behind glass windshields causing reflections to exist in the captured scene. This leads to poor understanding of the surrounding environment. Furthermore, removing reflections is an important pre-processing step for many video processing applications and systems. For example, one of the most common tasks in video surveillance applications is classifying and tracking objects. Reflections greatly degrade the localization and tracking accuracy of such algorithms. Robust systems for scene understanding, e.g., robots and self-driving cars, require removing reflections to factor out the noise in their visual representations and increase their accuracy. 

A video containing reflections can be viewed as a combination of two layers: background layer and reflection layer.  Thus, removing reflection artifacts from the input video is equivalent to decomposing the video into two layers. This, however, is a challenging and ill-posed problem, as there could be infinite number of valid decompositions. Most current methods for reflection removal are targeted towards single images. Applying such methods on videos frame by frame results in temporal flickering and incomplete separation. A recent work on videos \cite{nandoriya2017video} utilizes temporal information to overcome the problem of temporal flickering. However, this work assumes that the relative motion of the two layers is non-dynamic and easily distinctive. This assumption leads to incomplete separation in many natural scenes where the motions in the different layers are complex and dynamic. 

We propose a method to remove reflections from videos with complex motion and reflection patterns. 
The proposed method incorporates simple user hints with the temporal information available naturally in videos using a computational approach. Our method uses motion cues to separate the background and reflection layers. Sparse user annotations (hints) are used to improve the layer separation, especially in videos with complex motions. We have implemented the proposed method and compared it against the most recent video reflection removal method in \cite{nandoriya2017video} as well as the state-of-the-art image reflection method in \cite{fan2017generic}, after extending it to support video sequences. To our best knowledge, there are no datasets available publicly for videos with reflections. Thus, we captured and collected videos from prior works in various natural scenarios to test the proposed method in different conditions. The dataset has videos shot indoors, outdoors, in mobile environments, and on different reflective surfaces. (We will make our dataset public after the review process.) Natural scenes with reflections have no ground truth decomposition. Thus, we created synthetic videos that mimic the behavior of reflections to provide quantitative analysis between our method and prior works. Our performance analysis shows that our method significantly improves the separation output measured both qualitatively and quantitatively.

The rest of this paper is organized as follows. We summarize the related work in Section~\ref{sec:relatedWork}. We formally define the video reflection removal problem and present our solution for it in Section~\ref{sec:solution}. We describe our experimental evaluation in Section~\ref{sec:evaluation}, and we conclude the paper in Section~\ref{sec:conclusion}.


\begin{figure*}[htp] 
  \centering
  \includegraphics[width=\textwidth]{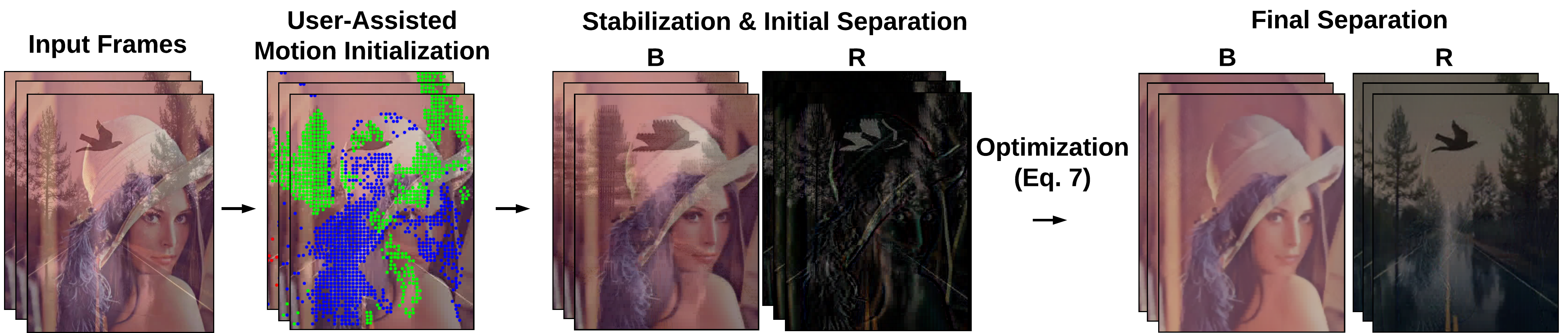}
  \caption{Overview of the proposed method. We start by estimating the motion fields for each layer with the help of sparse user annotations. We then utilize the motion fields to stabilize each layer and provide initial estimates for the background and reflection frames. Lastly, the initial estimates pass through an optimisation step to produce the final separated outputs.}
  \label{blockdiagram}
\end{figure*}

\section{Related Work} \label{sec:relatedWork}

The problem of removing reflections has been explored extensively in the image domain under several setups. Whereas the problem of video reflection removal has received less attention; it has been addressed only in \cite{nandoriya2017video}. We review the related literature on image as well as video reflection removal.

Reflections in natural images are a particular case of layer composition, where two layers are mixed together through addition forming the final image. Traditional single image reflection removal handles the ill-posed nature of the problem by relying on strong assumptions. For example, Levin et al. \cite{levin2003learning} use statistics of derivative filters and edge detectors in natural sequences as image priors to decompose the image into two layers. Li and Brown  \cite{li2014single} assume that both the background and the reflection layers have sparse gradients where the latter is much smoother. Shih et al. \cite{shih2015reflection} utilize sparse image statistics and encode them as a Gaussian mixture model. They further constrain the solution using the ghosting artifact assumption. Recently, Arvanitopoulos et al. \cite{arvanitopoulos2017single} suppress reflection artifacts by constraining the number of non zero gradients on the output. However, these assumptions can only cover a limited number of natural scenarios.

 Fan et al. \cite{fan2017generic} recently explored image reflection removal using deep neural networks. This work introduces two sub-networks: one for predicting the edge map of the background layer and another to reconstruct the background layer by adopting this edge map. However, this technique is limited only to blurring reflection artifacts. In scenarios with strongly textured reflections, the edge prediction sub-network fails and ruins the  background layer reconstruction. In \cite{wan2018crrn}, gradients reconstructions are used as hints to recover the details of the background. In \cite{zhang2018single}, perceptual and adversarial losses are used to recover the visual perception and properties of the separated images. A recent work uses a non-linear synthesis model with generative adversarial network (GAN) \cite{wen2019single} to remove reflections from single images.

All the previously mentioned approaches are designed to remove reflections from one image. In this paper, we focus on recovering the background and reflection layer in a video sequence. Simply extending the previous techniques of images to videos, such as applying the method on a frame by frame manner, does not provide accurate results as it leads to incomplete separation and temporal flickering as observed in \cite{nandoriya2017video}. 
To overcome these issues, Nandoriya et al. \cite{nandoriya2017video} proposed an extension of the work in \cite{xue2015computational} by formulating an initialization and optimization strategy taking into account the temporal aspects to remove reflections from videos.  This has shown to overcome the temporal flickering issue. However, this approach assumes that the two layers have simple non-dynamic motion that is easily distinctive for each layer. 
For example, if two objects in one layer move with different speeds and directions, the algorithm assigns each object to a different layer instead of both objects to the same layer, failing to provide an accurate separation.  Furthermore, their work utilizes motion trackers that fail to track weak features with low color information often found in natural sequences with reflections. For example, if a reflection part is blurry with low color information, their method fails to track this feature and thus it remains in the recovered background layer. 

In this work, we develop a new method that can handle common reflections without relying on specific motion constraints that do not work on the general case of reflections. We overcome the limitations associated with motion estimation by incorporating minimal user assistance. Our method shows good separation results in videos with complex reflection texture and dynamic motion.

\section{Problem Definition and Proposed Method} \label{sec:solution}

\subsection{Problem Definition}
We mathematically model a video with reflections as a composition of two layers as shown in Eq.~(\ref{eqn:framecomp}), where $I_t$ is the video frame, $B_t$ is the background layer and $R_t$ is the reflection layer at time $t$.
\begin{equation}
I_t = B_t \hspace{3pt} + \hspace{3pt} R_t.   
\label{eqn:framecomp}
\end{equation}

Making use of the temporal relationship between frames in videos, we define a motion field $W^{B}_{t,\rho}$ as the background layer warping motion field from frame $\rho$ to frame $t$. Similarly, $W^{R}_{t,\rho}$ is defined as the reflection layer warping motion field from frame $\rho$ to frame $t$. Thus, the video compositing equation can be defined as:
\begin{equation}
I_t = W^{B}_{t,\rho} B_{\rho} + W^{R}_{t,\rho}  R_{\rho}, \;\;  t=1,2,\dots, N.
\label{eqn:videocomp}
\end{equation}
 
The problem addressed in this paper can be stated as follows: Given an input video of length $N$ frames that is a mixture of background and reflection layers, we would like to decompose this video into two separate layers each with $N$ frames. Clearly, there is an infinite number of valid decompositions which makes this problem severely ill-posed, and thus challenging to solve.

\subsection{Overview of the Proposed Method} \label{sec:method}

The idea of the proposed video reflection removal method is to use motion cues to separate the background layer from the reflection layer. In addition, since motion in natural videos are quite complex, we utilize sparse user hints in the layer separation. 

A high-level illustration of the proposed method is shown in Figure \ref{blockdiagram}. We start by estimating the motion of each layer with the help of sparse user hints. We utilize the estimated motion to stabilize both layers and provide initial separation. We then perform an optimization process on the initially separated layers to provide the final output for each layer.

As detailed in Section~\ref{sec:motioninit}, motion initialization consists of two parts, computing motion tracks and then clustering them. Motion initialization is actually quite difficult to perform, because motions of objects in the background and reflection layers can overlap and/or obstruct each other, which can lead to incomplete layer separation. Thus, we propose a user-assisted motion initialization method that improves the accuracy of the estimated motion tracks. This is done with the help of an intuitive graphical user interface that improves motion tracking by allowing the user to annotate new tracks and in turn improves the clustering accuracy of these tracks.

After motion initialization, motion fields are known for both layers. In Section \ref{sec:stab}, we describe how we use a sliding window and background motion fields to stabilize the background motion in this window. This results in a stable background layer that can be separated by temporal filtering. Thus, providing an initial estimate of the first background frame in this window. The corresponding reflection frame is then estimated as the residual. We shift the window frame by frame to estimate the initial layer separation across all frames.

Finally in Section \ref{sec:opt}, we describe the optimization process to improve the initial estimates and provide the final separated layers. The optimization function consists of three terms: data term, layer prior term, and a smoothness term. The data term is to make sure that the recovered layers satisfy the video compositing model in Eq. (\ref{eqn:videocomp}). The layer prior term imposes labeling constraints on the initially estimated layers. The smoothness term is used for the spatial smoothness of the recovered layers.

In the following subsections, we describe the details of each step in our method.

\begin{figure}[tp]
\centering
\includegraphics[width=\linewidth]{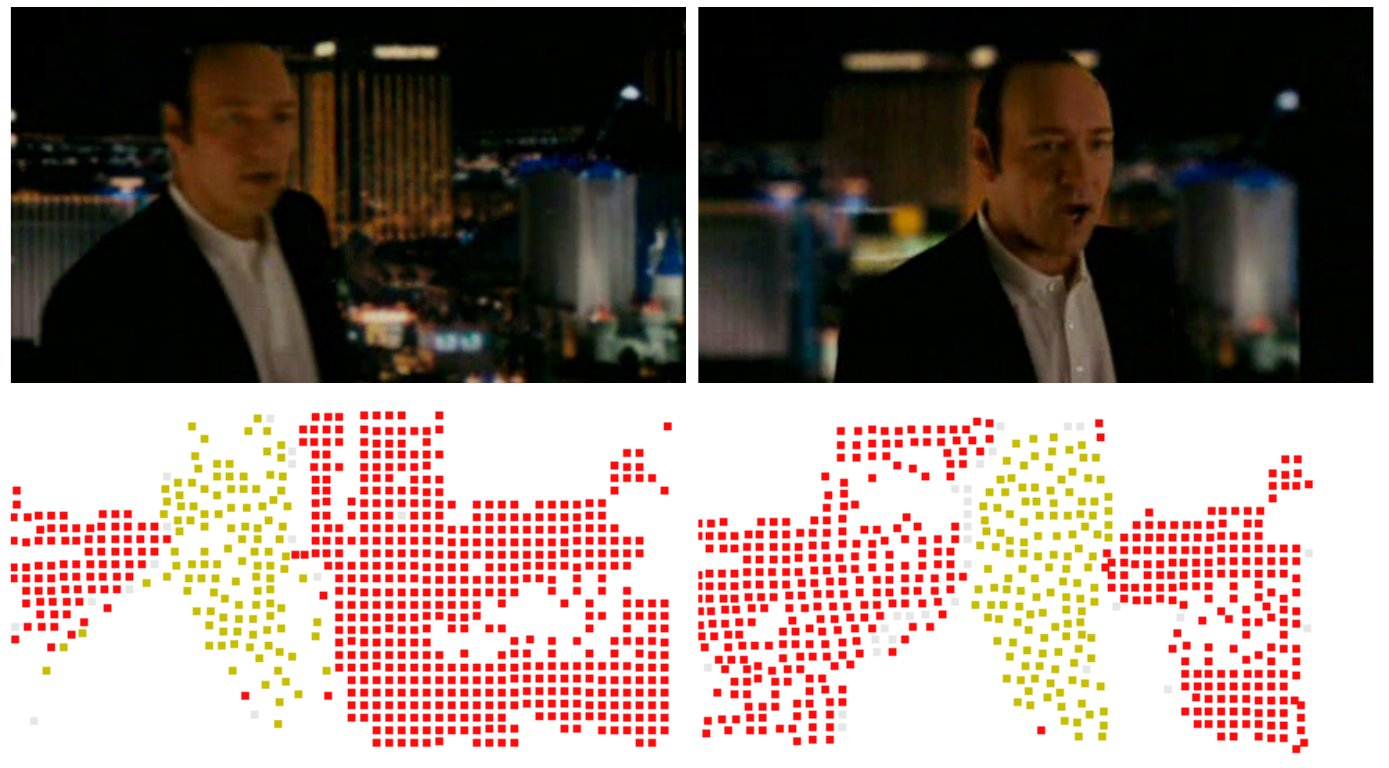}
\caption{Top: Two frames from a video shot, from the human motion database HMDB \cite{Kuehne11}. Bottom: Clustering of point tracks indicating regions with similar motion.}
\label{mosegscene}
\end{figure}

 \begin{figure}[tp]
  \centering
  \includegraphics[width=\linewidth]{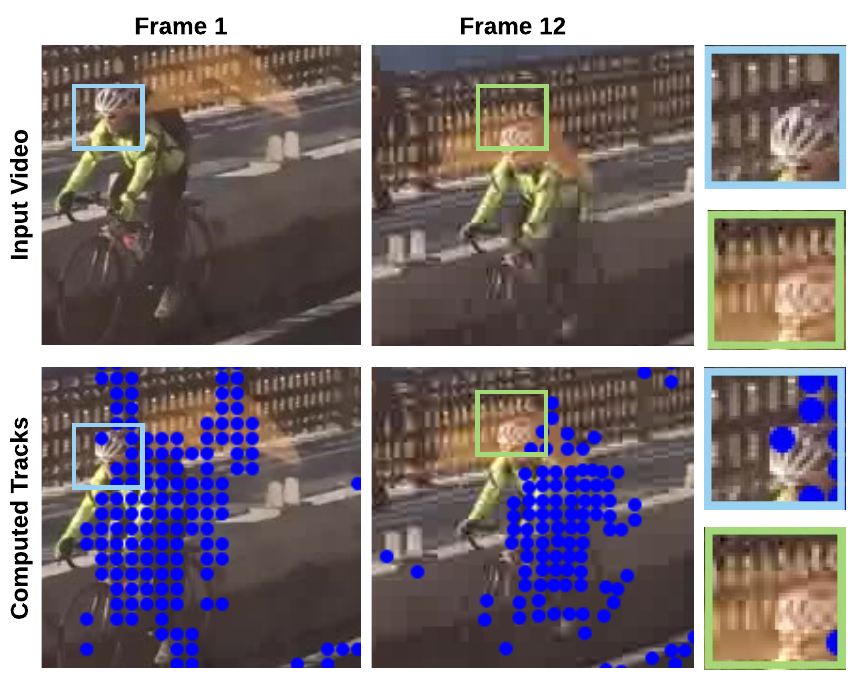}
  \caption{An example of dead tracks in videos with reflections. The helmet of the bicycle driver was obstructed by the reflection layer causing the track to die at frame 12.}
 \label{deadtrack}
 \end{figure}

 \begin{figure}[tp]
  \centering
  \includegraphics[width=1\linewidth]{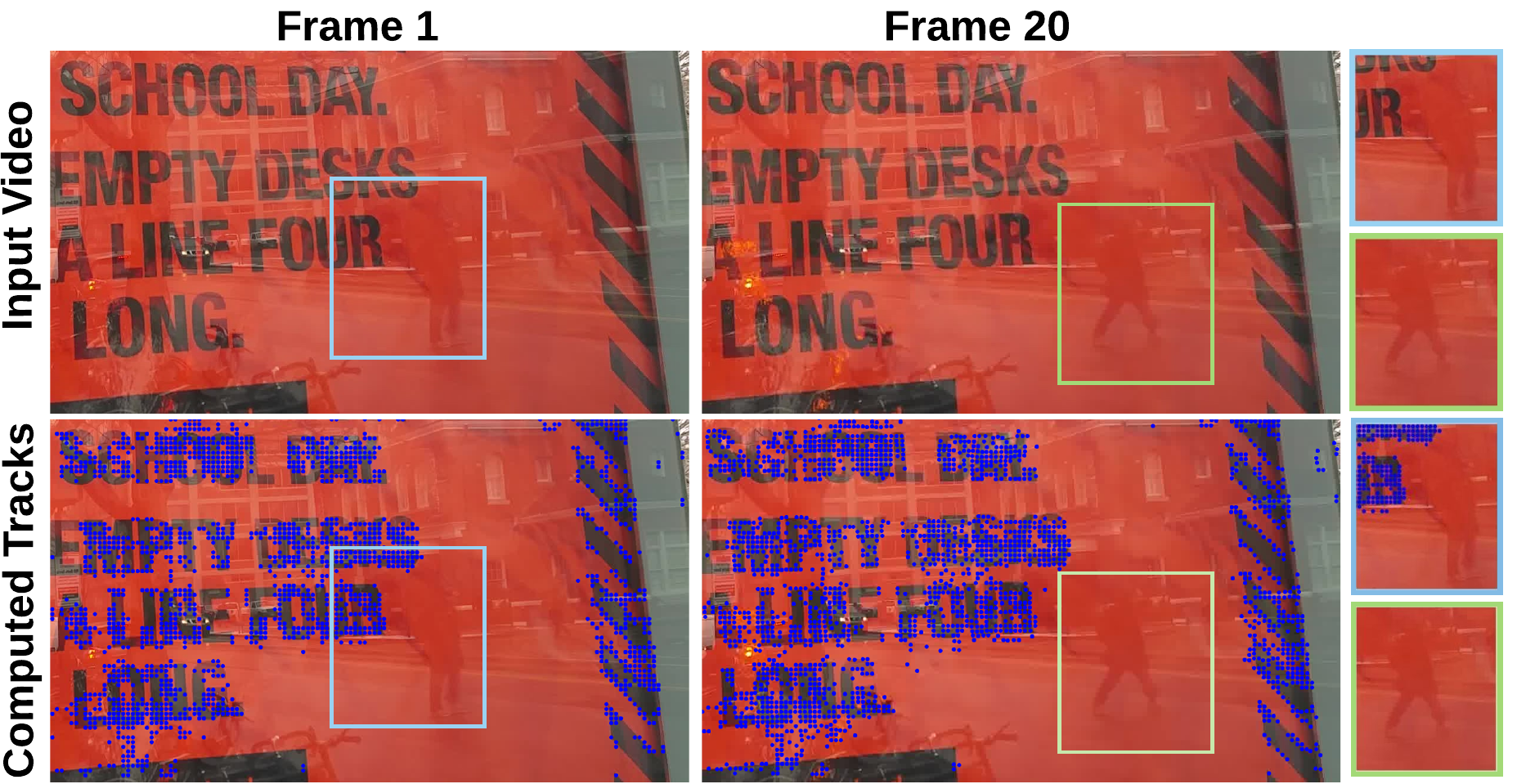}
  \caption{An example of weak features in videos with reflections. The tracker cannot capture the movement of the person in the reflection layer. Thus, when zooming on the boxes in the bottom row, no tracks (blue dots) are found.}
 \label{weakfeature}
 \end{figure}

\subsection{User-assisted Motion Initialization} \label{sec:motioninit}

The objective of the user-assisted motion initialization is to estimate the two dominating motions in the video. We achieve this by computing motion tracks and clustering them. 
Specifically, motion tracks have been shown before to help in classifying objects with different motions, e.g., in \cite{ochs2013segmentation}. To illustrate this concept, we show an example in Figure~\ref{mosegscene}. In this figure, a track is represented by a colored dot and defined as a point in the 2-dimensional space of a scene tracked over a number of frames. The figure shows two frames from a video sequence where the person in the scene has a different motion than the background. This is shown by the clustering of the green dots relative to the red dots. Thus, motion tracks over a sequence of frames can, in principle, differentiate among objects with different motions. 

As described before, a video with reflection is a mixture of two layers, where each layer is likely to have a motion pattern independent of the other layer. For example, when capturing a scene from inside a moving bus through a glass window, objects inside the bus will be reflected on the captured outside scene. Objects inside the bus, however, typically have different motion patterns than the outside objects. We exploit this observation to separate the background layer from the reflection layer.  In particular, the first step of our motion initialization is to compute motion tracks across multiple frames using a point-based tracking method such as \cite{sundaram2010dense}. This method tracks features such as corners and edges and provides subpixel accuracy. We chose point-based tracking instead of object-based tracking, e.g., \cite{vondrick2013efficiently}, because it provides finer granularity and hence it can potentially improve the accuracy of layer separation in videos with reflection. 

Motion tracking, however, faces two main problems when applied on videos with reflection, namely: dead tracks and weak features. A dead track occurs when a trajectory is identified for a limited number of frames only. One of the most common reasons that dead tracks exist in videos with reflections is occlusion and/or obstruction by the other layer. This makes the tracker not able to recognize the features of the target layer in the frames where the obstruction occurs from the other layer, and thus the trajectory dies. We illustrate the problem of dead tracks in Figure~\ref{deadtrack}, where the helmet of the person on the bike is detected in frame 1 but due to the obstruction from reflection its trajectory died in frame 12. 
 
The second problem for motion tracking in videos with reflection is the frequent presence of weak features, i.e, features having low contrast that are difficult to detect by the tracker. Color information is essential for the tracker to detect and track a feature, and if there is not enough color information, target tracking is hard or not possible. We show an example of weak features in Figure~\ref{weakfeature}, where the motion tracker could not capture the movement of the person in the reflection layer due to the low color information.
 
Dead tracks and weak features make motion tracking in videos with reflection very challenging, and therefore results in inaccurate layer separation and visual distortions. To address this problem, we propose using sparse user hints to guide the reflection removal process. Specifically, we utilize user hints in two ways: (i) first to improve the labeling of the preliminary tracks and (ii) second to add more tracks. Preliminary tracks are the ones automatically computed by the point tracker. We design a simple graphical user interface to collect sparse hints from users in one frame. Then, we carefully propagate these hints within the frame as well as to other frames. Our graphical user interface is based on the motion clustering method in \cite{shankar2015video}, which utilizes temporal propagation, long term motion, color distributions, and volume consistencies. 

\begin{figure}[tp]
\centering
\includegraphics[width=\linewidth]{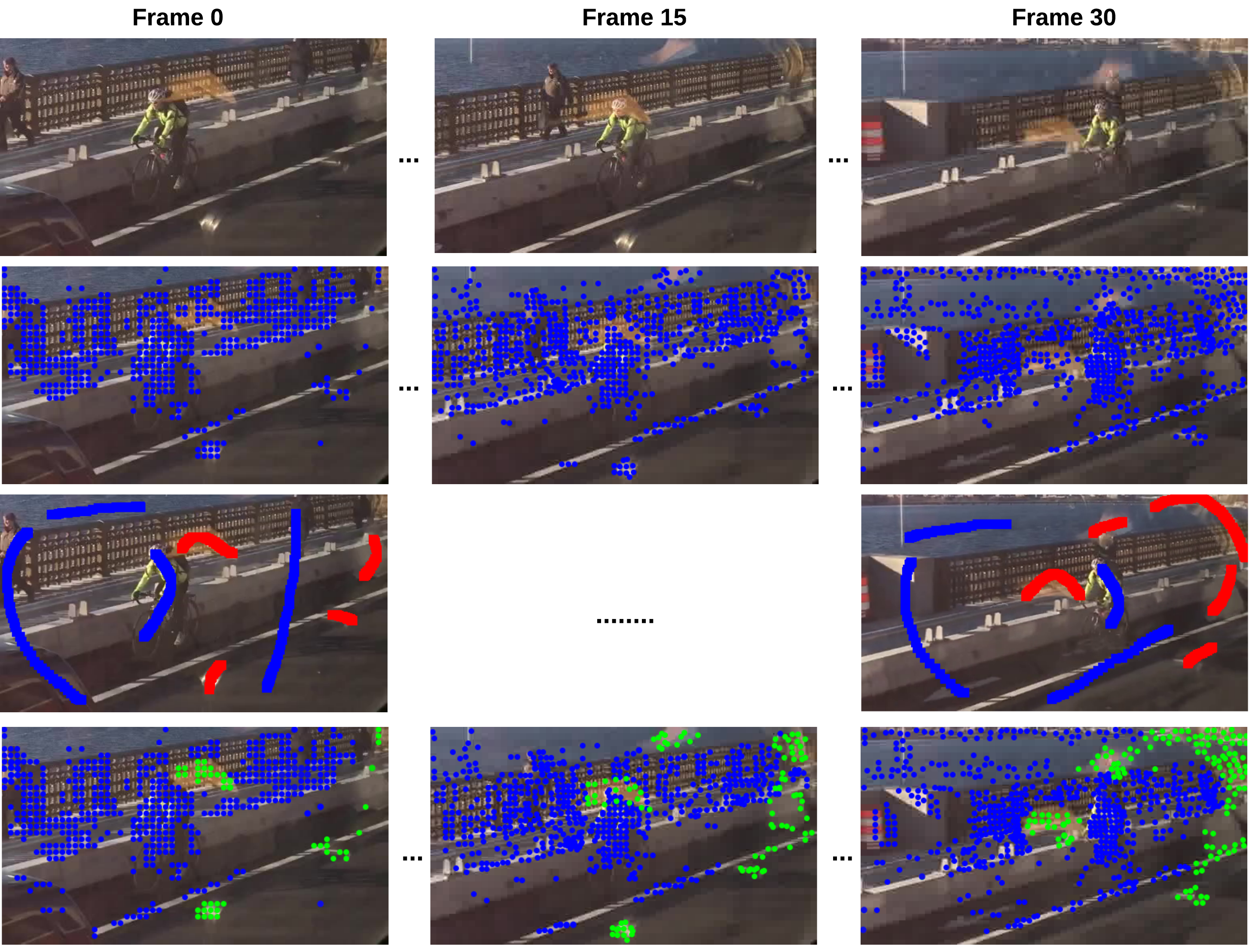}
\caption{Illustration of spares user assistance in motion initialization. The first row shows frame samples of the given video with reflections. The preliminary tracks are shown in the second row. The user sparsely provides simple annotations as shown in the third row. The blue and red hints correspond to the background and reflection areas, respectively. The annotations are propagated within the frame and across the frames as shown in the bottom row.}
\label{clusteringguiori}
\end{figure}

We illustrative our user-assisted motion initialization method by the example in  Figure~\ref{clusteringguiori}. The top row in the figure shows samples of the video from frame $0$ to frame $30$. The second row shows samples of the preliminary unlabeled tracks. In the third row, we show the spares user hints in frame $0$. The hints are given in form of blue and red scribbles referring to the background and reflection, respectively. These hints propagate the labeling to all other tracks in the same frame using a random walk computation \cite{grady2006random}. This means that within one frame while annotating the video, the user does not have to mark all areas, just make a few scribbles in areas with and without ref election. Then, the labeling is propagated across frames within the considered window of frames using the point-based tracking method \cite{sundaram2010dense}. 
We combine the newly added tracks by the user with the preliminary tracks. Each track $t_i$ is identified by a set of $(x_{ij},y_{ij})$ coordinates and by a label $l_i$. The $(x_{ij},y_{ij})$ coordinate indicates the spatial position of track $t_i$ in frame $j$. The label $l_i$ indicates to which layer this track belongs to; 1 for the background and 2 for the reflection. Using the labeled tracks, we obtain the initial motion fields for each layer between each pair of frames. To calculate the motion fields from frame $i$ to frame $j$, we first identify which tracks in frame $i$ are still tracked at frame $j$. However, since many tracks will die from frame $i$ to frame $j$ if the difference between them is large, we create a sliding window to secure more point correspondence between each pair of frames. We choose $10$ as the length of our window as it showed good separation in most results. Then, in each window, we calculate the projective warping matrix (homography) from each frame to the first frame in this window using the shared point tracks between them. This is done using the iterative re-weighted least squares (IRLS) similar to \cite{zaragoza2013projective}. Then, we slide the window frame by frame and calculate the projective warping matrix between each frame and the first frame in this window.

\subsection{Stabilization and Initial Layer Separation} \label{sec:stab}

After motion initialization, the projective warping matrices are known for both the background and reflection. We use the background warping matrices to stabilize the background layer and provide initial background estimation. To perform this, we use a sliding window of length $10$ frames. We stabilize the background motion in this window by warping the last $9$ frames in it onto the first one using the background warping matrices. This results in a warped version of each frame on the first frame in this window where the background motion is nearly stable. We then extract the initial background layer of the first frame in the window using minimum temporal filtering which calculates the minimum intensity across all warped frames. We use a minimum operation instead of the mean since the minimum is an upper bound for the background’s intensity \cite{szeliski2000layer}. This is based on the idea that the reflection layer can only add to the intensity of the dominant (background) layer, i.e., any contribution from the reflection layer will only be increasing the background layer intensity. So for each pixel we calculate its minimum value in all the warped frames and assign this value to the recovered background layer.

Let $I^{B}_{r,i}$ be the warped version of frame $i$ on the reference frame $r$, i.e., the first frame in the window, using the estimated background warping field $W^{B}_{r,i}$. The initial estimate of the background layer frame $r$ of the sliding window $\Omega$ can then be calculated as:

\begin{equation}
\widehat{B_r} = min(\widehat{B_r}, I^{B}_{r,i}),  \hspace{5pt} \forall i \in \Omega.
\label{eqn:b1estimate}
\end{equation}

The corresponding reflection layer can be taken as the residual component from Eq. (\ref{eqn:framecomp}) as:
\begin{equation}
\widehat{R_r} = I_r - \widehat{B_r}.  
\label{eqn:riestimate}
\end{equation}

We then slide the window frame by frame and perform the same operation to get initial estimates of all the background and reflection frames.

\subsection{Optimization} \label{sec:opt}

Optimization is done on the initial estimates of the background and reflection frames to improve the separation accuracy. The optimization function consists of three main terms, \textcolor{black}{the data term ($E_d$), layer prior term ($E_l$) and the smoothness term ($E_s$) as shown in Eq. (\ref{eq:optimgeneral}).} $\lambda_d$, $\lambda_l$ and $\lambda_s$ are weights we assign to each of these terms respectively. In our experiments, $[2,2,1]$ for $[\lambda_d,\lambda_l,\lambda_s]$ has  shown to provide good results.
\begin{equation}
E = \lambda_d E_d  +  \lambda_l E_l + \lambda_s E_s. 
\label{eq:optimgeneral}
\end{equation}

The data term $E_d$ is to make sure that the recovered layers satisfy the video compositing model formulated earlier in Eq. (\ref{eqn:videocomp}). This is done by minimizing the error between a layer at time $t$ and its warped version from time $\rho$ as shown in Eq. (\ref{eq:dataterm}) where $\| x\|_1$ is the L1-norm and $N$ is the total number of frames in the video. \textcolor{black}{We chose the L1-norm for its robustness in optimization \cite{yang2016robust}. The linear additive model itself is simple and have been used successfully in the past in solving many computer vision problems including reflection \cite{li2014single}.}

\begin{equation}
E_d = \sum\limits_{t=1}^N\sum\limits_{\rho=1}^N \| B_t - W^{B}_{t,\rho}B_{\rho}\|_1 + \| R_t - W^{R}_{t,\rho}R_{\rho}\|_1.
\label{eq:dataterm}
\end{equation}

To provide some prior information to our optimisation we use a similar approach to \cite{li2013exploiting}. We define a layer prior term $E_l$ that imposes labeling constraints on the initial estimated layers. We do this by defining a binary map $M_t$ indicating to which layer each pixel belongs to. \textcolor{black}{ $M_t$ = 0 for the background, and $M_t$ = 1 otherwise.This is based on the assumption that the background edges and the reflection edges are independent. That is, if we observe a strong gradient in the input image, it most likely belongs either to the background
component or the reflection component, but not to both \cite{xue2015computational}. As explained earlier during stabilization, we stabilize the motion of the background layer by warping all frames in the window on the first frame. Making the background motion in the warped (aligned) frames stable means that the high frequency components, such as edges, will have nearly stable magnitudes in all warped frames. However, the high frequency components not belonging to the background will have sparse magnitudes across the warped frames. We then estimate $M_t$ once by thresholding the alignment errors of the high frequency components during the background layer stabilization.} The formulation for the layer prior term is shown in Eq.~(\ref{eqn:layerprior}), where $\nabla I_t$ is estimated by a canny edge detector and $|\nabla B_t|$ and $|\nabla R_t|$ are the first order spatial gradients of the background and reflection respectively.
\begin{equation}
E_l = \sum\limits_{t=1}^N (M_t\nabla I_t |\nabla B_t| + (1-M_t)\nabla I_t |\nabla R_t|).
\label{eqn:layerprior}
\end{equation}

The third term $E_s$ in Eq. (\ref{eq:optimgeneral}) is used to provide spatial processing on the reconstructed frames by enforcing spatial smoothness. \textcolor{black}{ This is done by minimizing the first order spatial gradients as follows:}
\begin{equation}
E_s = \sum\limits_{t=1}^N (|\nabla B_t| + |\nabla R_t|).
\end{equation}

\section{Evaluation} \label{sec:evaluation}

In this section, we first describe the video dataset used in the experiments. Then, we assess the performance of our method on several videos. Then, we compare our methods against two recent methods in the literature. Then, we compare our method as well as others using a synthetically created video for which we can compute the ground truth reflection layer.  Finally, we analyze the impact of the user-provided hints on the performance of our method. 

\subsection{Dataset}

Our dataset consists of roughly 156 videos. The number of frames in each video ranges between 62 to 120 frames. We shot some of the videos and collected others from prior work by \cite{xue2015computational} and \cite{nandoriya2017video}. This dataset will be made public after the review process. 

\begin{figure}[tp]
  \centering
  \includegraphics[width=\linewidth]{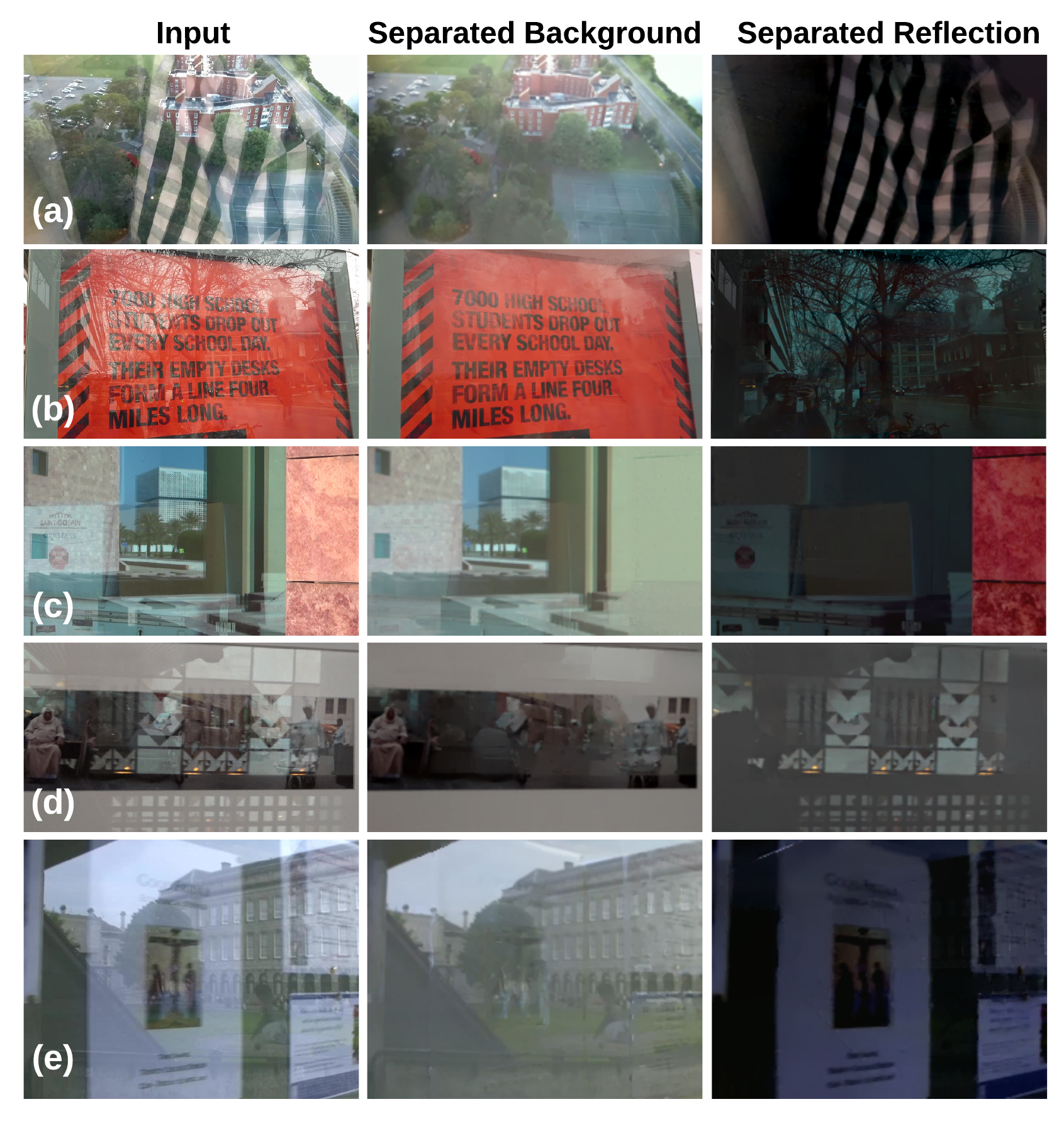}
  \caption{Example outputs of our method on natural videos.}
\label{fig:manyresults}
\end{figure}

We have performed experiments on real and synthetic videos from the dataset under different scenarios, with different background and reflection objects and various lighting conditions. Some videos were taken indoors (Figure~\ref{fig:manyresults}(a) and (d)) and outdoors (Figure~\ref{fig:manyresults}(b)). Videos were taken at different times of the day to represent different lighting conditions. For example, some videos were taken during twilight (figure is not shown due to space limitations) and others in the afternoon (Figure~\ref{fig:oriwithandwithout}). Some videos (Figure~\ref{fig:manyresults}(a), (b), and (c)) have highly textured background and reflections which our method manages to recover accurately. 


\subsection{Performance of our Method}

We first note that the best way to see our results is by viewing the actual videos with and without reflection. We have created a combined sequence with multiple videos showing the performance of our method as well as others, \textbf{which is posted (anonymously) on YouTube at: \url{https://youtu.be/7mnyB9-J-vY}}. 

In the following, we only present sample representative frames from multiple sequences, due to space limitations. 

Figure \ref{fig:manyresults} shows the result of our method on a few sample videos where reflections frequently occur. One of these common cases is capturing reflective surfaces such as glass covered billboard (Figure~\ref{fig:manyresults} (b), (d)). Our method performs well in such sequences. For example, in Figure \ref{fig:manyresults} (b), the reflection is strong and highly textured with many details of the reflection objects, e.g., the trees and the person capturing the video. However, our method manages to produce good separation where words on the sign are clearly visible and the reflection details are recovered. 

Moreover, reflections often occur when capturing outdoor scenes through indoor windows as shown in Figures~\ref{fig:manyresults} (a), (c) and (e). Our method produces a clean recovery of the outdoor scene while showing the indoor scene details that were hard to see in the original video. Recovering the background is usually the subject of interest, however, recovering details from the reflection scene might be useful in some cases where information needs to be extracted from the reflection layer.

\begin{figure*}[tp]
  \centering
  \includegraphics[width=1\linewidth]{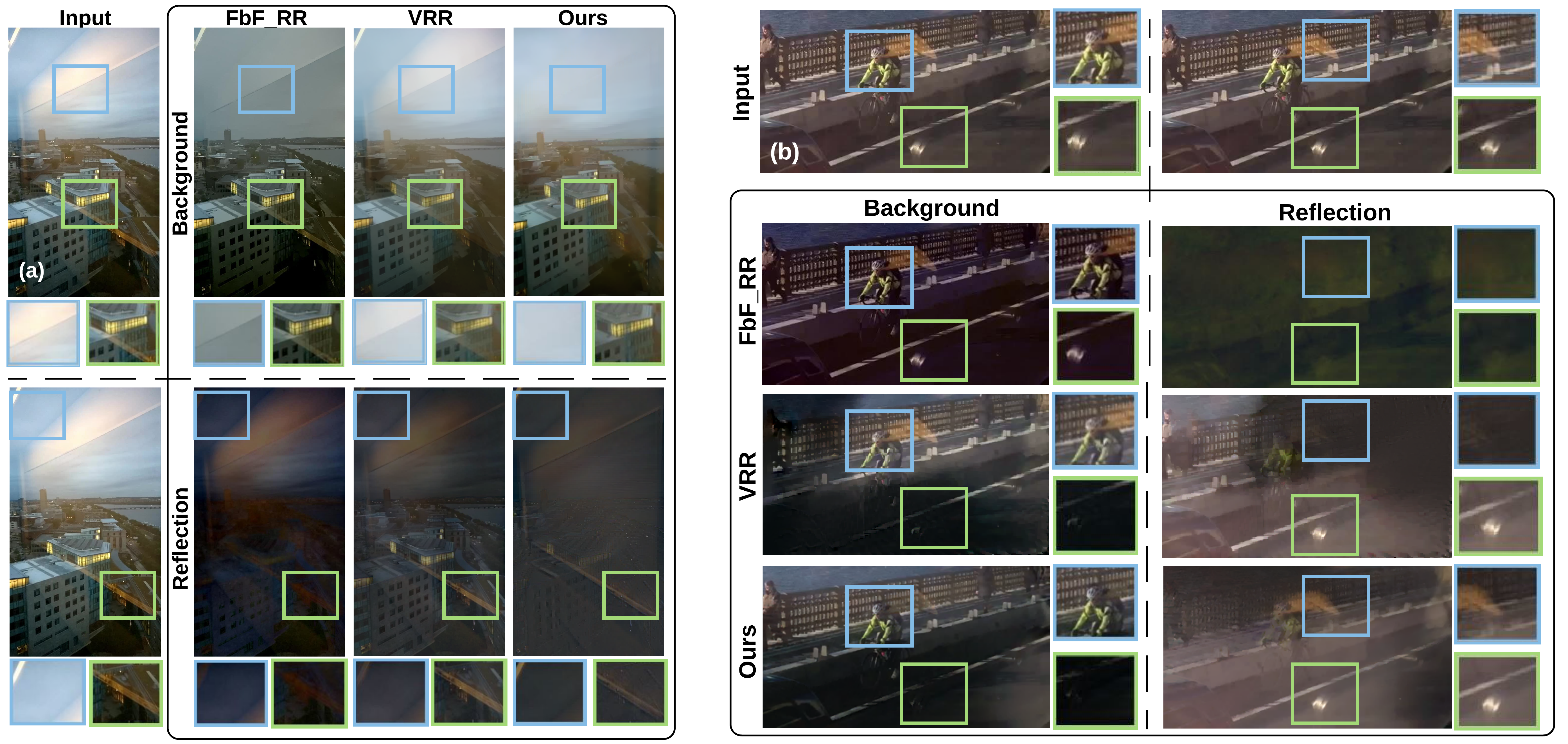}
  \caption{Comparison of our method versus state-of-the-art on two natural videos with reflections. We zoom in with blue and green patches on different locations in the background and reflection layers to show the differences.}
  \label{fig:ajayfanours}
\end{figure*}

\subsection{Comparison Against Stat-of-the-Art}

We compare our method against two state-of-the-art methods for removing reflections. To the best of our knowledge, there is only one work specifically designed for removing reflections in videos in \cite{nandoriya2017video}. We compare against this work and the work done by Fan et al. \cite{fan2017generic} for removing reflections in images.
To compare against \cite{nandoriya2017video}, we used the code provided by the authors and refer to this method as VRR (short for Video Reflection Removal) in the figures. To compare against \cite{fan2017generic}, we used the code publicly available and extended it to work on videos by applying their method on a frame by frame manner. We refer to this work as FbF\_RR. 

In Figure~\ref{fig:ajayfanours}, we show side-by-side comparisons on two sequences. In the first sequence, shown in Figure \ref{fig:ajayfanours} (a), our method manages to remove reflections from the background layer completely, while there are noticeable artifacts in the results of the other approaches. For example, as shown in the blue and green patches in the background layer, FbF\_RR results in a blurry background with noticeable color changes and reflection artifacts. VRR leads to incomplete separation and noticeable reflection artifacts. Our method produces a clean background layer with no noticeable reflection effects and better color consistency. Moreover, the reflection image recovered by our method is cleaner than the ones produced by the other methods. For example, as shown in the green patch in the reflection layer, the buildings in the background are still visible in the recovered reflection layer in the results of VRR, while the results of FbF\_RR have color changes and unclear reflection recovery. Our method recovers the reflection part in this area more accurately. In the blue patch, FbF\_RR cannot capture the reflection part in the corner, VRR results in a blurred recovery while our approach recovers most of the reflection texture. 

In Figure~\ref{fig:ajayfanours} (b), we show a challenging sequence where the video was taken from inside a moving bus. The background includes various objects with complicated local movements. For the background layer, other methods result in color changes and reflection artifacts, while our method produces clean separation with most of the reflection parts removed. For example, the blue patch on the helmet of the bicycle driver still has the reflection part in VRR and FbF\_RR results, while our method removes most of the reflection part in this area. In the green patch, FbF\_RR cannot remove the reflection part while both our approach and VRR manage to remove it. Since this is a challenging sequence, the reflection part recovery is hard. However, our approach recovers most of the reflection parts. As shown in the blue patch, FbF\_RR and VRR cannot recover the reflection structure in this area while our method manages to recover most of the reflection texture. 

In summary, the proposed method significantly improves layer separation compared to current methods in the literature.

\begin{figure}[tp]
\centering
\includegraphics[width=\linewidth]{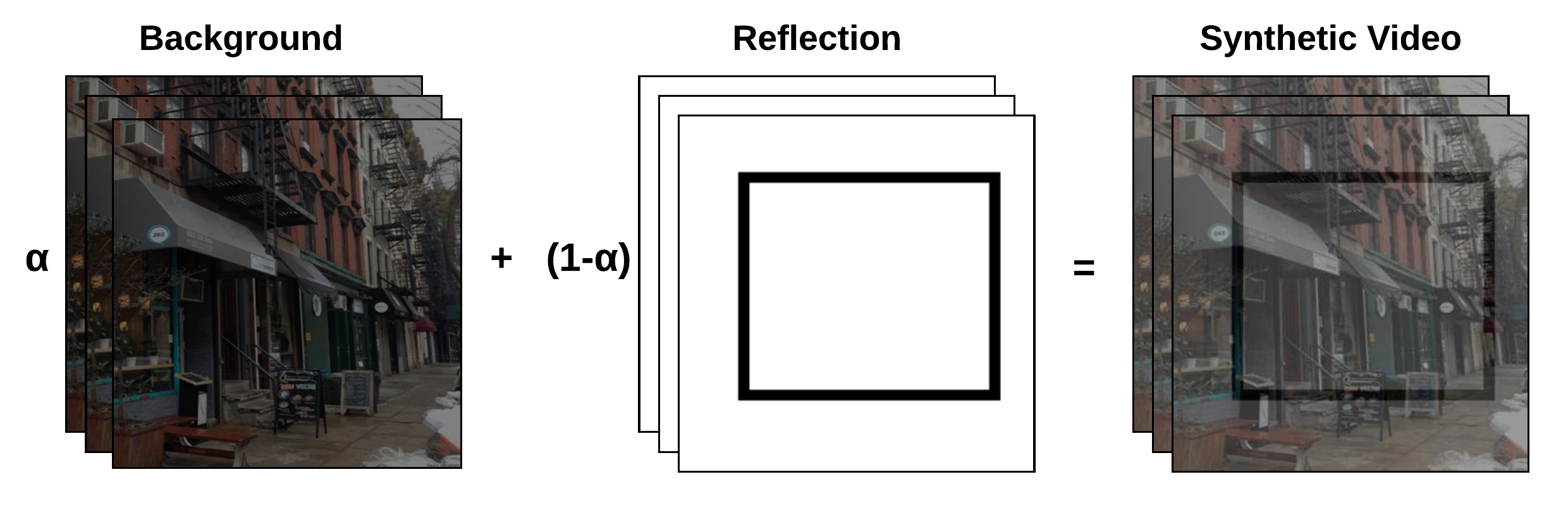}
\caption{Creating a synthetic video representing the ground truth to objectively evaluate the proposed video reflection removal method.}
\label{fig:synth}
\end{figure}

\begin{figure} [tp]
  \centering
  \includegraphics[width=\linewidth]{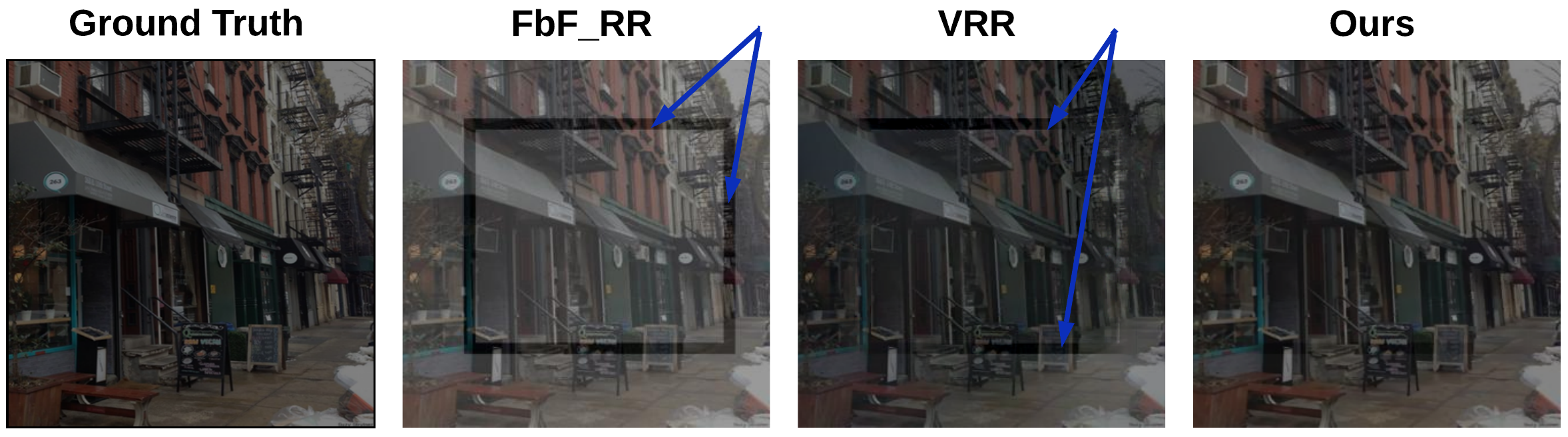}
    \caption{Comparison of our method against state-of-the-art using a ground truth (synthetic) video.}
\label{fig:synth_results}
\end{figure}

\begin{figure}[tp]
\centering
\subfloat[][Background]
{\includegraphics[width=0.48\linewidth]{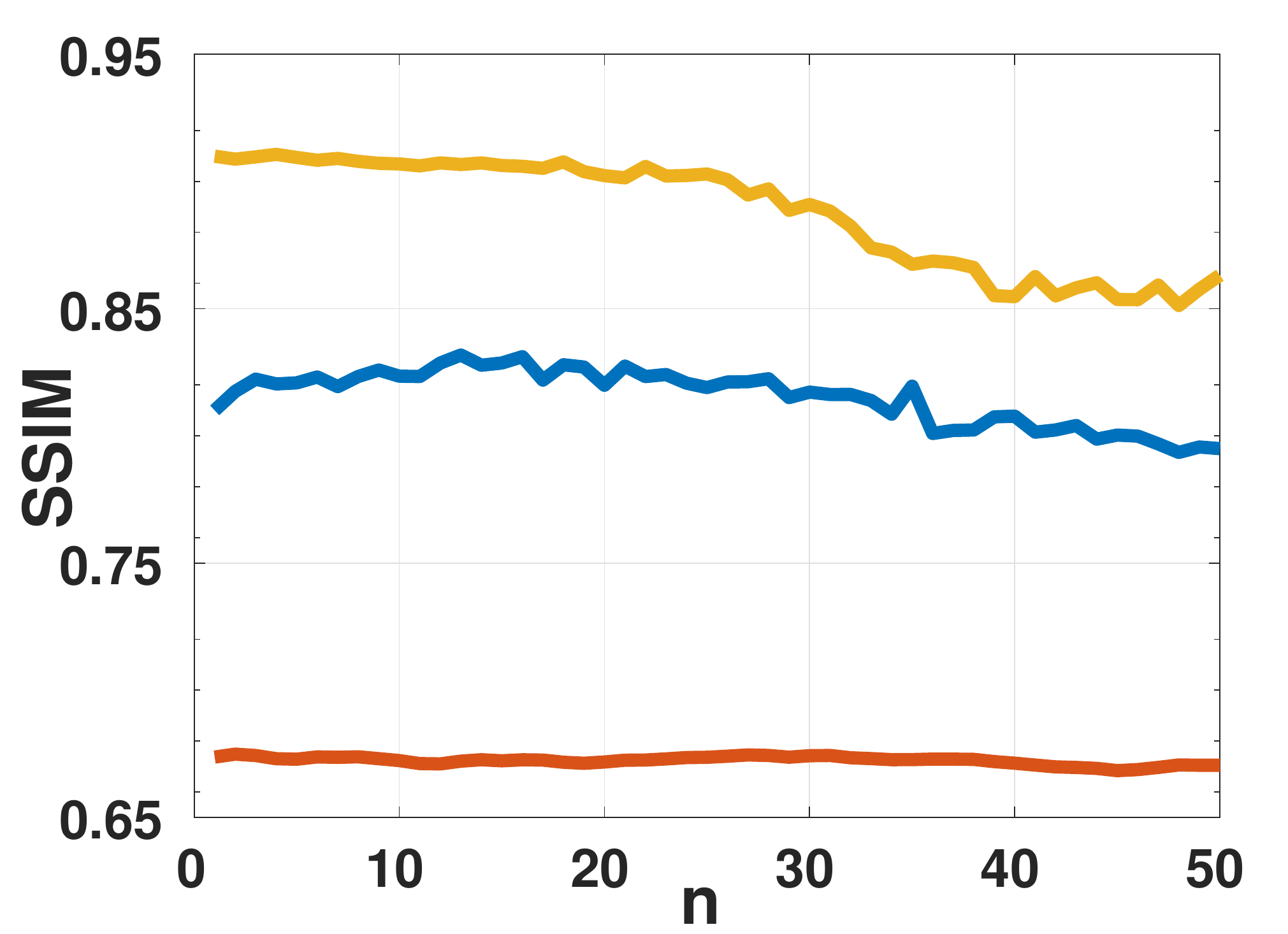}
 \label{fig:sfig1}}
\subfloat[][Reflection]
{\includegraphics[width=0.48\linewidth]{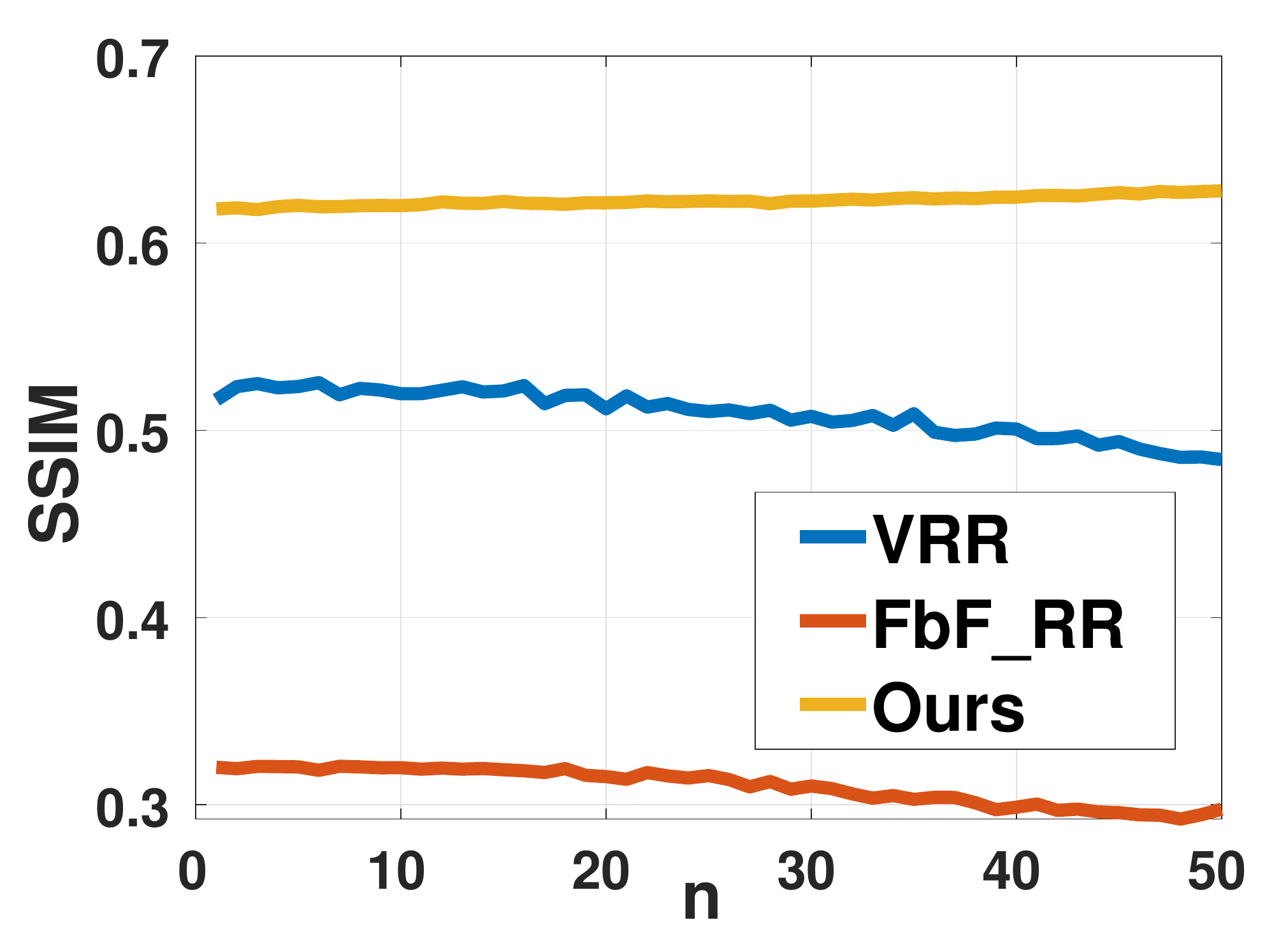}
 \label{fig:sfig2}}
\caption{SSIM measured against the ground truth for the background and reflection frames. $n$ is the frame number.}
\label{fig:ssim}
\end{figure}

\begin{figure*}[tp]
  \centering
  \includegraphics[width=\linewidth]{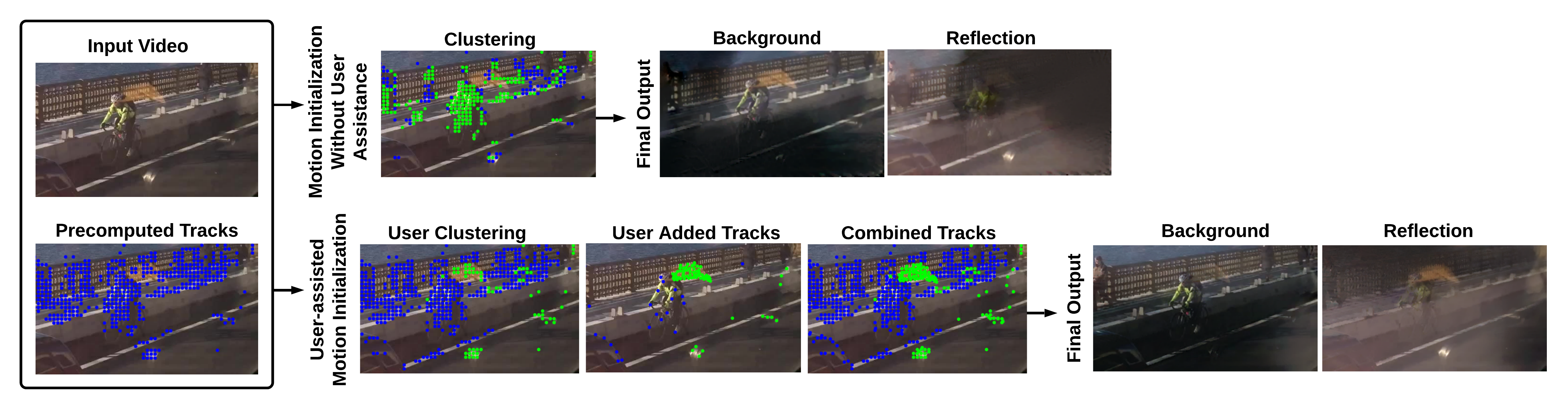}
  \caption{Impact of user-assistance. User-assistance leads to more accurate motion initialization and final separation.}
\label{fig:oriwithandwithout}
\end{figure*}

\subsection{Comparison Against Ground Truth}

We are not aware of any public dataset of videos with reflections along with their ground truth layer separation. It is, actually, impractical to capture such ground truth separations in a natural environment, as this would need capturing the same scene with and without the reflective surface at the same time. 

Recall that a video with reflection is modeled as two mixed layers. To mimic this scenario, we generate synthetic videos using an additive layer composition model. We do this by averaging two synthetic videos $V_1$ and $V_2$ with different coefficients as $V = \alpha V_1 + (1-\alpha) V_2$, 
where $\alpha$ is the mixing parameter ranging between 0 and 1. For example, Figure \ref{fig:synth} shows the output of such generation scheme for $\alpha = 0.8$.
In this video, we treat the buildings picture as the background and the black square as the reflection. We introduce two different global movements for both layers of magnitude +3 and -3 pixels per frame in the horizontal direction for the background and reflection layers,  respectively.

After generating the synthetic sequence, we process it using FbF\_RR, VRR, and our method. The background reconstruction results for this sequence are shown in Figure~\ref{fig:synth_results}. The figure shows that that our method produces more accurate reflection removal than others for the synthetic example. For instance, FbF\_RR background reconstruction was blurry and has reflection artifacts. VRR results were visually better, but still included reflection artifacts. Our method produced a cleaner background reconstruction than the other methods, where reflection is not visible in the reconstructed background layer.

Finally, we objectively compare our methods versus FbF\_RR and VRR, by calculating the accuracy of separation using the spatial similarity of the recovered layers with the ground truth. Spatial similarity is measured using the Structural Similarity Index (SSIM) \cite{wang2004image}. An SSIM value of 1 means perfect match between the video layer (background or reflection) and the corresponding ground truth, i.e., perfect separation.  
Figure \ref{fig:ssim} shows the values of the SSIM measured against the ground truth for both the background and reflection frames. The figure shows that our method consistently outperforms the other two methods.

\subsection{Impact of User Assistance}
Finally, we analyze the importance of user-assistance in our approach. We show the results of applying our method with and without user assistance on a sample sequence in Figure~\ref{fig:oriwithandwithout}. The same preliminary tracks are used in both cases. Clustering without user assistance is performed using a k-means clustering. We show the motion initialization output in both cases. The figure shows that motion initialization is more accurate with user hints, which results in better final layer separation.

\section{Conclusion} \label{sec:conclusion}

We have presented a user-assisted method to remove undesired reflections from videos. The method utilizes motion cues as well as sparse user hints for separating the background layer from the reflection layer. We overcome the limitations associated with the state-of-the-art methods for reflection removal by improving the separation results of videos with a complex dynamic motion that is hardly distinctive for each layer. We presented quantitative and qualitative results on challenging real and synthetic examples. Our method produces clean separation of both the background and reflection layers. We compared against state-of-the-art video reflection removal and video extensions of image reflection removal methods. Our quantitative and qualitative results show that our method leads to significant improvements in the separation quality than prior works. 


\balance

\clearpage
\bibliographystyle{ACM-Reference-Format}
\balance
\bibliography{main}

\end{document}